\documentclass[twoside,11pt]{article}

\usepackage{blindtext}

\usepackage[abbrvbib, preprint]{jmlr2e}
\usepackage{xcolor} 
\usepackage{amsmath}
\usepackage{amssymb}
\usepackage{mwe}
\usepackage{multirow}
\usepackage{booktabs}

\usepackage{lastpage}
\jmlrheading{23}{2026}{1-\pageref{LastPage}}{3/26}{9/22}{21-0000}{Lai and Simnacher}

\ShortHeadings{CF for two-sample testing}{Lai and Simnacher}
\firstpageno{1}

\begin{document}

\title{Counterfactual Explanations for Deep Two-Sample Testing}

\author{\name Wei-Cheng Lai \email wei-cheng.lai@hpi.de \\
       \addr Digital Health and Machine Learning\\
       Hasso-Plattner-Institute, University of Potsdam\\
       Potsdam, Germany
       \AND
       \name Marco Simnacher \email marco.simnacher@hu-berlin.de \\
       \addr Chair of Statistics\\
       Humboldt-Universität zu Berlin\\
       Berlin, Germany
       \AND
       \name Christoph Lippert \email christoph.lippert@hpi.de \\
       \addr Digital Health and Machine Learning\\
       Hasso-Plattner-Institute, University of Potsdam\\
       Potsdam, Germany\\
       \addr Hasso Plattner Institute for Digital Health at Mount Sinai\\
       Icahn School of Medicine at Mount Sinai\\
       New York, United States of America}

\editor{My editor}

\maketitle

\begin{abstract}
Two-sample testing is a fundamental tool for detecting distributional differences across scientific domains, but classical tests (including kernel-based tests) can be ineffective on high-dimensional structured data such as images. Recent deep two-sample tests improve sensitivity in these settings by learning informative representations, yet they provide limited insight into which data features drive rejection of the null hypothesis $H_0$.
To address this issue, we propose a counterfactual explanation framework for deep two-sample testing that generates sample-level edits moving observations from a source group toward a target group while explicitly reducing the discrepancy measured by the test. Our method combines a diffusion autoencoder with a pretrained deep two-sample test model and optimizes a maximum mean discrepancy (MMD) objective in the test model's representation space to produce plausible counterfactuals. We quantify distribution-level effects through changes in the test statistic and the resulting two-sample p-values.
We evaluate the method on synthetic 2D shape datasets and two MRI cohorts. Across both settings, the counterfactual transformations consistently increase p-values relative to the original samples, indicating that the edited source set becomes statistically closer to the target distribution under the test. We measure minimality using LPIPS to ensure the counterfactuals remain close to the original samples. The resulting edits provide interpretable evidence of the features associated with the detected group differences. On MRI, the localized changes are consistent with known anatomical differences between cohorts.
\end{abstract}


\section{Introduction}
Two-sample tests assess whether two sets of observations arise from the same underlying distribution and are foundational in medicine, biology, psychology, and the social sciences, where the goal is to detect population-level differences between groups. 
They are typically formulated with a null hypothesis and an alternative hypothesis. The null hypothesis assumes that two sets of observations come from the same underlying distribution, while the alternative hypothesis states that they come from different distributions.

While classical tests lose power on high-dimensional and structured data such as images \cite{JMLR:v13:gretton12a}, deep two-sample tests using deep learning architectures learn better latent representations and achieve powerful statistics on complex data \cite{Kirchler2020,lopez-paz2017revisiting}. In these methods, structured data are first mapped to latent representations using deep neural networks, and a statistical test is then applied in the learned feature space. 
While this substantially improves performance on complex data, the output of the test typically remains limited to a test statistic, a $p$-value, or a reject/non-reject decision. As a result, deep two-sample tests provide little insight into which semantic features drive the detected distributional difference.

This lack of interpretability is particularly limiting in medical imaging, where researchers and clinicians need to understand how cohorts differ and whether the detected differences correspond to clinically meaningful patterns \cite{xai_medical}. 
Existing explainable AI methods mainly focus on supervised prediction models. Gradient-based methods highlight influential input regions, while counterfactual methods explain predictions through minimal input changes. However, these approaches are typically designed for classifiers and do not directly explain group differences captured by a two-sample test statistic.

\begin{figure}[t]
    \centering
    \includegraphics[width=\textwidth, height=8cm, keepaspectratio]{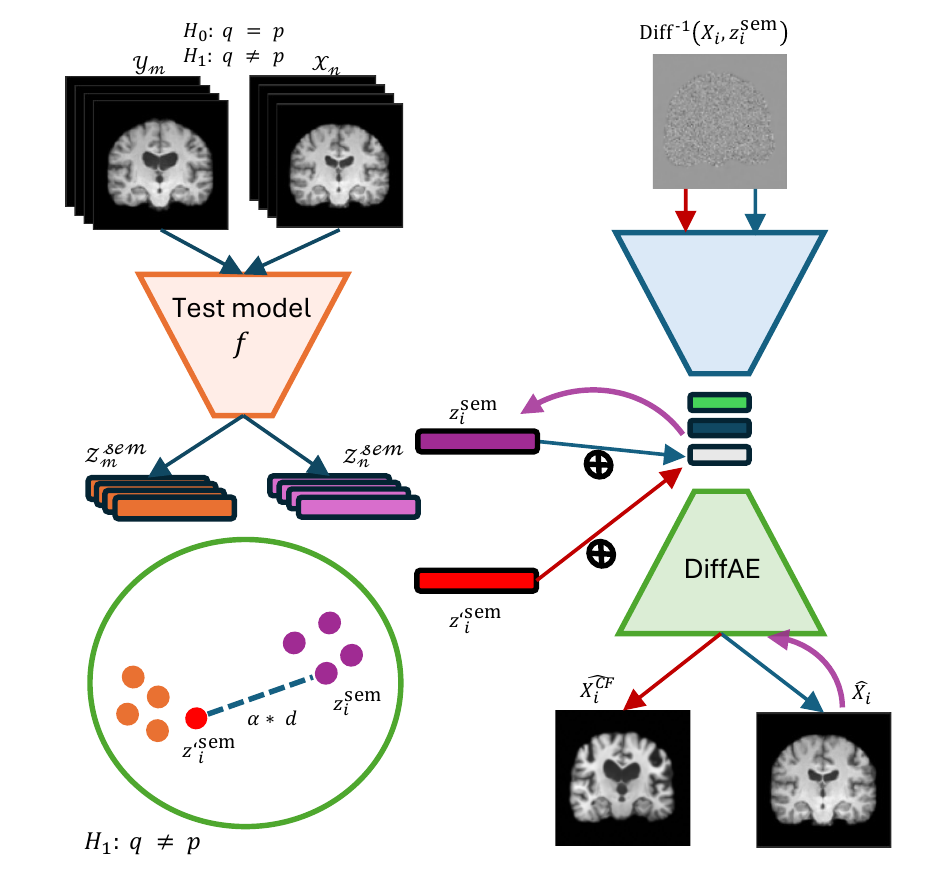}
    \caption{Overview of the proposed deep-test-guided counterfactual generation framework. A pretrained deep two-sample test $f$ defines a discrepancy objective between source and target groups. A source sample $X_i$ is encoded in the latent space of a diffusion autoencoder $z^{sem}_i$ and edited along a direction $d$ given by the gradient of this objective, moving it toward the target distribution $\mathcal{Y}_m$. The resulting counterfactual provides a sample-level explanation of the detected group difference.}
    \label{fig1_framework}
\end{figure}

In this work, we propose a counterfactual explanation framework for deep two-sample testing, illustrated in Fig.~\ref{fig1_framework}. Given a source sample and a target group, our method generates a plausible edit that moves the source sample toward the target distribution while explicitly reducing the discrepancy measured by the test. We formulate this as a counterfactual editing problem in the latent space of a diffusion autoencoder, guided by a pretrained deep two-sample test model \cite{Kirchler2020}. This yields explanations at both the sample level, through interpretable visual edits, and the group level, through changes in the test statistic and $p$-value.

Our goal is to answer the following questions: which features drive the separation between two groups, and what minimal changes are required to make samples from one group statistically closer to the other? Therefore, we make the following contributions: First, we introduce a counterfactual explanation framework for deep two-sample testing. Second, we generate plausible edits by optimizing a maximum mean discrepancy-based objective in the latent space of a diffusion autoencoder guided by a pretrained test model. Third, we evaluate the resulting explanations on synthetic and MRI data using both visual and statistical criteria.

\section{Related Work}

Deep two-sample tests improve classical two-sample testing on high-dimensional data by learning representations in which group differences can be detected more effectively \cite{lopez-paz2017revisiting,Kirchler2020}. However, these methods typically provide only a test statistic or decision, without explaining which semantic features drive the detected difference.

In explainable AI, gradient-based methods provide post-hoc saliency maps that highlight input regions relevant to a model output \cite{grad_input,gradcam,guided_backprop,integrated_grad}. Another line of work uses counterfactual explanations, which identify input changes that alter a model prediction \cite{counterfactual_singla,gifsplanation_cohen,preechakul2021diffusion}. While effective for supervised models, these approaches are generally designed for classifiers rather than statistical tests over groups.

Our work connects these directions by introducing counterfactual explanations for deep two-sample testing. Instead of explaining a class prediction, we explain group-level distributional differences by generating plausible edits that reduce the discrepancy measured by the test.

\section{Background}
\subsection{Deep Two-sample test}
Two-sample tests are used to detect whether two sets of data are from the same distribution. For high-dimensional data, many studies embed the data onto lower-dimensional representations and apply two-sample tests to these representations  \cite{Kirchler2020,liu2020learning,tian2024unified}.
While our framework and the corresponding counterfactual explanations are generally applicable to such two-sample tests, from here on, we focus on the deep two-sample test of ~\cite{Kirchler2020}, since it performed well on high-dimensional, structured image data by embedding it through a deep statistical test model.

Let $p$ and $q$ be probability measures on $\mathbb{R}^d$ with a common dominating Borel measure. The deep two-sample test tests $H_0:q=p$ against $H_1:q\neq p$ based on two samples drawn from $p$ and $q$. 
Therefore, it assumes to have access to four samples $\mathcal{X}_n, \mathcal{Y}_m, \mathcal{X}'_{n'}, \mathcal{Y}'_{m'}$ with observations drawn independently from $p,q,q'$ and $p'$, respectively, where $p'$ and $q'$ are assumed to be similar to $p$ and $q$, respectively. In particular, we have $n$ observations $X_i\overset{iid}{\sim} p$ for sample $\mathcal{X}_n=\{X_1,\hdots, X_n\}\subset \mathbb{R}^d$ (analogous notation for $\mathcal{Y}_m, \mathcal{X}'_{n'}, \mathcal{Y}'_{m'}$). 

In the first step, the deep two-sample test uses the auxiliary samples $\mathcal{X}'_{n'}$ and $\mathcal{Y}'_{m'}$ to learn the deep statistical test model $f:\mathbb{R}^d\to \mathbb{R}^H$ to minimize the distance between the average embeddings of $\mathcal{X}'_{n'}$ and $\mathcal{Y}'_{m'}$.  The embedding model can be e.g. a ReLU network mapping inputs onto a layer of dimension $H$ of a neural network \cite{Kirchler2020}. In the second step, $f$ is fixed and on the test samples $\mathcal{X}_n$ and $\mathcal{Y}_m$, the mean distance between the embeddings is computed as
\begin{equation}
    D_{n,m}(f)= \overline{f(\mathcal{X}_n)} - \overline{f(\mathcal{Y}_m)} 
    \ = \frac{1}{N}\sum_{i=1}^Nf(X_i) - \frac{1}{M}\sum_{i=1}^Mf(Y_i)
\label{eq:mean_distance}
\end{equation}
where $\overline{f(\mathcal{X}_n)}=\frac{1}{n}\sum_{i=1}^nf(X_i)$.
Then, the test statistic called the Deep Maximum Mean Discrepancy (DMMD)  is defined as 
\begin{equation}
    S_{n,m}(f, \mathcal{X}_n,\mathcal{Y}_m) = \frac{n m}{n + m} \left\lVert D_{n,m}\right\rVert^2
    \label{eq:dmmd}
\end{equation}
Then, p-values can be obtained by comparing the observed test statistic to test statistics computed with permuted group labels \cite{ernst2004permutation}.

\subsection{Diffusion Autoencoder}
Learning compact and semantically meaningful representations for downstream analysis is less straightforward in standard diffusion-based models (DPMs) \cite{ho_ddpm,improved_ddpm-nichol21a} than in explicitly latent-variable models such as VAEs, GANs, Latent diffusion models (LDMs) and Diffusion Autoencoder (DiffAE) \cite{KingmaW13_vae,NIPS2014_gan,Rombach_2022_ldm,preechakul2021diffusion}.

To generate realistic counterfactual images, we use a diffusion autoencoder (DiffAE) \cite{preechakul2021diffusion}. DiffAE maps an input image $X_i \in \mathbb{R}^d$ to a semantic latent representation $z_i^{\mathrm{sem}} \in \mathbb{R}^H$ through an encoder
\[
f_{\mathrm{sem}}:\mathbb{R}^d \to \mathbb{R}^H.
\]
A diffusion-based decoder then reconstructs the image from the semantic code and a stochastic noise variable $X_T$:
\[
\hat{X}_i = \mathrm{Diff}(X_T, z_i^{\mathrm{sem}}).
\]
DiffAE is suitable for our setting because it enables counterfactual editing in a semantic latent space while preserving realistic image synthesis.


\subsection{Visual Counterfactual Explanations}

Visual counterfactual explanations (VCEs) explain model predictions by generating an edited image that changes the model output while remaining similar to the original sample \cite{counterfactual_singla,Singla2020Explanation}. For a classifier $\phi$, a standard counterfactual objective ~\cite{globalCF} takes the form
\[
L_{\mathrm{CF}}(\hat{X}^{CF}_i)=\phi(\hat{X}^{CF}_i)+\lambda\, s(X_i,\hat{X}^{CF}_i),
\]
where $\hat{X}^{CF}_i$ denotes the counterfactual image and $s(X_i,\hat{X}^{CF}_i)$ measures similarity between the original and edited image, commonly using LPIPS (Learned Perceptual Image Patch Similarity) \cite{zhang2018_lpips}. The first term encourages a change in model output, while the second enforces minimality between the counterfactual and original images. In Section~3, we adapt this idea from classifier explanations to explanations of group differences under a deep two-sample test.
Global Counterfactual Directions (GCD)~\cite{globalCF} extends visual counterfactual explanations with the DiffAE framework~\cite{preechakul2021diffusion}. Given the semantic latent representation $z^{\mathrm{sem}}_i \in \mathbb{R}^H$ of an image, the method searches for a direction in latent space that minimizes a counterfactual objective. The modified latent code $z'^{\mathrm{sem}}_i$ is then decoded to generate the counterfactual image $\hat{X}^{CF}_i$.


\section{Counterfactual Explanations for Deep Two-Sample Tests}

Existing counterfactual explanation methods are mainly designed to explain classifier decisions \cite{globalCF,counterfactual_singla}. In contrast, our goal is to explain group-level differences detected by a deep two-sample test. When the null hypothesis $H_0$ is rejected, the key question is not which changes would flip a class label, but which semantic changes would make a source sample statistically closer to the target group under the test statistic. We address this problem by combining the deep two-sample test of \cite{Kirchler2020} with latent counterfactual editing based on GCD \cite{globalCF}.

\subsection{Proposed counterfactual explanation for two-sample tests}
\label{sec:cf_inference}
Let the source and target observations be denoted by $\mathcal{X}_n$ and $\mathcal{Y}_m$, respectively. Our goal is to generate a counterfactual version of a source sample $X_i \in \mathcal{X}_n$ such that the edited sample remains similar to the original image while making the source group less distinguishable from the target group $\mathcal{Y}_m$ under the deep two-sample test.

This idea follows naturally from the DMMD statistic in Eq.~\eqref{eq:dmmd}, which is based on the mean distance between source and target representations in Eq.~\eqref{eq:mean_distance}. If we perturb a source sample to obtain a counterfactual $\hat{X}^{CF}_i$ and replace $X_i$ by its counterfactual $\hat{X}^{CF}_i$ in the source set, then reducing this mean distance also reduces the test discrepancy between the modified source observations and the target observations.

We therefore define a counterfactual explanation for two-sample testing as a minimal edit to a source sample that reduces the discrepancy between the source and target groups while preserving similarity to the original image. This differs from classifier-based counterfactual explanations, whose objective is typically to change a predicted class label \cite{globalCF}.

We rewrite the counterfactual loss as follows. Let $D_{n,m}(\cdot)$ denote the test discrepancy, let $f$ be the pretrained auxiliary test model, and let $\hat{X}^{CF}_i$ denote the counterfactual image corresponding to source sample $X_i$. We define the general counterfactual objective for a statistical test as
\begin{equation}
    L_{cf}(\hat{X}^{CF}_i) = D(\hat{X}^{CF}_i, \mathcal{X}_{n\backslash i}, \mathcal{Y}_m) + \lambda \, s(X_i, \hat{X}^{CF}_i),
    \label{eq:cf_twosample}
\end{equation}
where $\mathcal{X}_{n\backslash i}=\mathcal{X}_n \backslash X_i$, and $s(X_i,\hat{X}^{CF}_i)$ measures similarity between the original and edited image.

In this work, we instantiate the discrepancy term with the deep maximum mean discrepancy (DMMD) from Eq.~\eqref{eq:dmmd}. This yields the objective
\begin{equation}
    L_{cf}(\hat{X}^{CF}_i) = \left\lVert D_{n,m} \right\rVert^2 + \lambda \, s(X_i,\hat{X}^{CF}_i),
    \label{eq:cf_dmmd}
\end{equation}
which can be written explicitly as
\begin{equation}
    \hspace{-0.25cm}
    L_{cf}(\hat{X}^{CF}_i) =
    \left\lVert \overline{f(\mathcal{X}'_n)} - \overline{f(\mathcal{Y}_m)} \right\rVert^2
    + \lambda \, s(X_i,\hat{X}^{CF}_i),
    \quad
    \mathcal{X}'_n = \{\hat{X}^{CF}_i, \mathcal{X}_{n\backslash i}\} \subset \mathbb{R}^d.
    \label{eq:cf_mean_distance}
\end{equation}

Minimizing Eq.~\eqref{eq:cf_mean_distance} therefore corresponds to editing a single source sample such that the modified source set $\mathcal{X}'_n$ becomes semantically closer to the target observations $\mathcal{Y}_m$ in the learned representation space, while the edited sample $\hat{X}^{CF}_i$ remains close to the original image through the similarity weighted by $\lambda$.

Although the framework is not restricted to diffusion-based generators, in this work, we follow \cite{preechakul2021diffusion,globalCF} and use DiffAE to generate realistic counterfactual images.

\subsection{Finding counterfactual explanations with latent perturbations}

When the deep two-sample test rejects the null hypothesis $H_0:p=q$, the source observations $\mathcal{X}_n$ and target observations $\mathcal{Y}_m$ are inferred to come from different underlying distributions. To generate a counterfactual explanation of this difference, we train a diffusion autoencoder (DiffAE) and use the pretrained auxiliary test model $f$ as the semantic encoder.

During training, the parameters of $f$ are kept fixed, and the model is used only to extract semantic representations $z^{\mathrm{sem}}_i$ from the input images. Following \cite{preechakul2021diffusion}, the DiffAE is trained with the standard denoising objective on noisy latent representations.

At inference time, we optimize the objective in Eq.~\eqref{eq:cf_mean_distance} by perturbing the semantic latent representation of a source sample. For a source image $X_i$, we compute the gradient of the counterfactual loss with respect to its semantic code,
\[
d_i = \frac{\partial L_{cf}(\hat{X}_i)}{\partial z^{\mathrm{sem}}_i},
\]
and update the latent representation as
\[
z'^{\mathrm{sem}}_i = z^{\mathrm{sem}}_i - \alpha d_i,
\]
where $\alpha$ denotes the step size. The modified latent code $z'^{\mathrm{sem}}_i$ is then decoded by the trained DiffAE decoder to obtain the counterfactual image
\[
\hat{X}^{CF}_i = \mathrm{Diff}(X_T, z'^{\mathrm{sem}}_i).
\]
In this way, the generated counterfactual remains close to the original sample while reducing the discrepancy between the modified source set and the target set under the deep two-sample test. The resulting image, therefore, provides a visual explanation of which semantic changes move the source sample toward the target distribution.

\section{Experiments}

We evaluate the proposed method on two settings: structural T1-weighted brain MRI and synthetic shape data. The goal is to assess whether the generated counterfactuals provide meaningful explanations of group differences detected by a deep two-sample test.

We assess counterfactual quality along three complementary axes. First, \emph{statistical effectiveness} measures whether the generated counterfactuals move source samples closer to the target group under the downstream two-sample test, as quantified by changes in the test statistic (DMMD) in Eq.~\eqref{eq:dmmd} and in the corresponding $p$-value. Second, \emph{minimality} measures how much the counterfactual deviates from the source sample, using LPIPS, since a useful counterfactual should remain close to the original image. Third, \emph{realism} measures whether the generated samples remain plausible instances of the target distribution.

Because the proposed method is explicitly optimized to reduce the discrepancy measured by the downstream deep two-sample test, we treat the statistical criterion as the primary model-selection metric. Counterfactual images are generated by optimizing Eq.~\eqref{eq:cf_mean_distance}, where $\lambda$ controls the trade-off between distribution alignment and similarity to the original image.

\subsection{Brain MRI}

ADNI (Alzheimer's Disease Neuroimaging Initiative) \cite{adni_Petersen2010-zh} is a clinical dementia study with multiple clinical covariates, including CDR (Clinical Dementia Rating), as well as imaging-derived variables such as ventricular volume. In contrast, UK Biobank (UKBB) \cite{UKB} is a large population cohort containing health records and brain MRI from the general population.

We evaluate the proposed method on structural T1-weighted brain MRI to test whether the generated counterfactuals capture meaningful anatomical differences between clinically relevant groups. We use ADNI \cite{adni_Petersen2010-zh} for counterfactual evaluation and a pretrained deep two-sample test model trained on UKBB to guide counterfactual generation in the semantic latent space.

For each setting, we compare the proposed method against a conditional baseline counterfactual method and evaluate the generated samples using statistical effectiveness, minimality, and realism. Statistical effectiveness is measured through changes in the downstream two-sample test statistic and $p$-value, minimality is measured using LPIPS relative to the source images, and realism is measured using Precision, Recall, Density, and Coverage (PRDC) against target samples.

\subsubsection{Experimental Setup}
We use the coronal middle slice of T1-weighted MRI scans aligned to the MNI atlas. The auxiliary statistical model $f(\cdot)$ is implemented with a ResNet50 backbone and trained on UKBB ($\mathcal{X}'_{n'}, \mathcal{Y}'_{m'}$) to predict lateral ventricular volume. Following \cite{Kirchler2020}, we use the feature output from the last convolutional layer with a \textit{tanh} activation as the representation for the deep two-sample test.

We then evaluate counterfactual explanations on ADNI by defining source and target groups ($\mathcal{X}_n, \mathcal{Y}_m$) according to two clinically relevant variables: Clinical Dementia Rating (CDR) and age. For CDR, the source cohort consists of healthy brains ($\mathrm{CDR}=0$) and the target cohort consists of cognitively impaired brains ($\mathrm{CDR} \geq 1$). For age, we split the cohort at the median age of 74, with younger subjects as the source group and older subjects as the target group. In both settings, the original two-sample test detects significant group differences. We train DiffAE on ADNI while keeping the pretrained model $f(\cdot)$ fixed, and compare the proposed method against a label-conditional DiffAE baseline, which replaces the semantic encoder with a linear layer.

\subsubsection{Experimental Results}
\paragraph{Statistical effectiveness}

\begin{table*}[t]
\centering
\caption{Statistical evaluation of MRI counterfactuals at $\lambda=4$. Larger $\Delta p$ and $\Delta t$ indicate stronger alignment with the target distribution after counterfactual generation. The reported numbers are averaged over 6 seeds. ``Non-scale'' means that the test statistic is reported without normalization by the number of samples; see Eq.~\eqref{eq:dmmd}.}
\label{tab:mri_stats_lambda4}
\begin{tabular}{llcc}
\toprule
Setting & Method & $\Delta p$ mean & $\Delta t$ mean (non-scale) \\
\midrule
\multirow{2}{*}{CDR}
& Conditional CF & 0.00013 & 55.91 \\
& Proposed CF & \textbf{0.00865} & \textbf{176.40} \\
\midrule
\multirow{2}{*}{Age}
& Conditional CF & -0.00147 & -128.53\\
& Proposed CF & \textbf{0.01279} & \textbf{164.29} \\
\bottomrule
\end{tabular}
\end{table*}
We generate counterfactuals using six random seeds for each $\lambda \in \{0.0625, 0.125, 0.25, 0.5, 1, 2, 4\}$, following \cite{globalCF}. We also vary the number of generated counterfactuals, using $n \in \{20, 50, 100, 200, 300\}$ for CDR and $n \in \{20, 50, 100, 200, 400\}$ for age. Across this grid, the strongest performance on the primary statistical metrics (DMMD and $p$-value) is obtained consistently at $\lambda=4$, which we therefore use for the main statistical comparison.

We define
\[
\Delta t = D(\mathcal{X}_n, \mathcal{Y}_m) - D(\hat{\mathcal{X}}^{CF}_n, \mathcal{Y}_m), 
\qquad
\hat{\mathcal{X}}^{CF}_n = \{\hat{X}^{CF}_1, \dots, \hat{X}^{CF}_{n}\},
\]
where a larger $\Delta t$ indicates that the discrepancy to the target cohort is reduced more strongly after editing. Likewise, $\Delta p$ denotes the change in $p$-value before and after counterfactual generation, where larger values indicate that the edited source set is statistically closer to the target set under the downstream test.

Table~\ref{tab:mri_stats_lambda4} shows that the proposed method outperforms the conditional baseline on both MRI settings at this operating point. In the CDR setting, the proposed method increases the mean $\Delta p$ from $1.3\times 10^{-4}$ to $8.65\times 10^{-3}$ and the mean $\Delta t$ from 55.91 to 176.40. In the Age setting, the baseline yields negative average shifts on both metrics, whereas the proposed method yields positive improvements. Overall, these results indicate that the proposed objective is substantially more effective at moving source samples toward the target distribution under the deep two-sample test.

\paragraph{Minimality and realism.}

\begin{table}[t]
\centering
\caption{Minimality evaluation of MRI counterfactuals at $\lambda=4$ using LPIPS between source and counterfactual images. Lower is better.}
\label{tab:mri_lpips_lambda4}
\begin{tabular}{llc}
\toprule
Setting & Method & LPIPS \\
\midrule
\multirow{2}{*}{CDR}
& Conditional CF & 0.1296 \\
& Proposed CF & \textbf{0.1157} \\
\midrule
\multirow{2}{*}{Age}
& Conditional CF & \textbf{0.0212} \\
& Proposed CF & 0.1196 \\
\bottomrule
\end{tabular}
\end{table}
Table~\ref{tab:mri_lpips_lambda4} reports LPIPS at $\lambda=4$, measuring perceptual similarity between the source and counterfactual MRIs. In the CDR setting, the proposed method yields lower LPIPS than the baseline (0.1157 vs.\ 0.1296), indicating more localized edits that remain closer to the source image. In the Age setting, however, the baseline is substantially better on LPIPS. This suggests that improved statistical alignment does not necessarily imply improved source-faithfulness across all cohort definitions.

\begin{table*}[t]
\centering
\caption{Realism evaluation of MRI counterfactuals at $\lambda=1$ using PRDC metrics between generated counterfactuals and target samples. Higher is better.}
\label{tab:mri_prdc_lambda1}
\begin{tabular}{llcccc}
\toprule
Setting & Method & Density & Coverage & Precision & Recall \\
\midrule
\multirow{2}{*}{CDR}
& Conditional CF & \textbf{0.3381} & \textbf{0.2901} & \textbf{0.3676} & \textbf{0.2554} \\
& Proposed CF & 0.3302 & 0.2733 & 0.3538 & 0.2449 \\
\midrule
\multirow{2}{*}{Age}
& Conditional CF & \textbf{0.4760} & \textbf{0.2484} & \textbf{0.4451} & \textbf{0.3345} \\
& Proposed CF & 0.3405 & 0.1789 & 0.4166 & 0.2493 \\
\bottomrule
\end{tabular}
\end{table*}

To complement the source-faithfulness analysis, Table~\ref{tab:mri_prdc_lambda1} reports PRDC realism metrics at $\lambda=1$, where realism is strongest or near-strongest. In the CDR setting, the proposed method remains close to the baseline across density, coverage, precision, and recall, suggesting that its statistical gains are not achieved at a major realism cost. In the Age setting, however, the baseline performs better on all PRDC metrics.

Taken together, these results show that the proposed objective is most convincing in the CDR setting, where it combines clear improvements on the primary statistical criterion with improved minimality and competitive realism. More broadly, the MRI experiments reveal a trade-off between statistical effectiveness, minimality, and realism across different cohort definitions.

Overall, the MRI results indicate that the proposed method is strongest when evaluated on the criterion it explicitly optimizes, namely discrepancy reduction under the downstream test. The auxiliary analyses further show that realism is optimized at a more moderate operating point, while minimality improves with larger $\lambda$, mainly in the CDR setting. This supports the interpretation that the LPIPS term acts as a useful regularizer, although its effect depends on the dataset and cohort definition.

\paragraph{Qualitative interpretation}
\begin{figure*}[t]
    \centering
    \begin{minipage}[t]{0.49\textwidth}
        \centering
        \includegraphics[width=\linewidth]{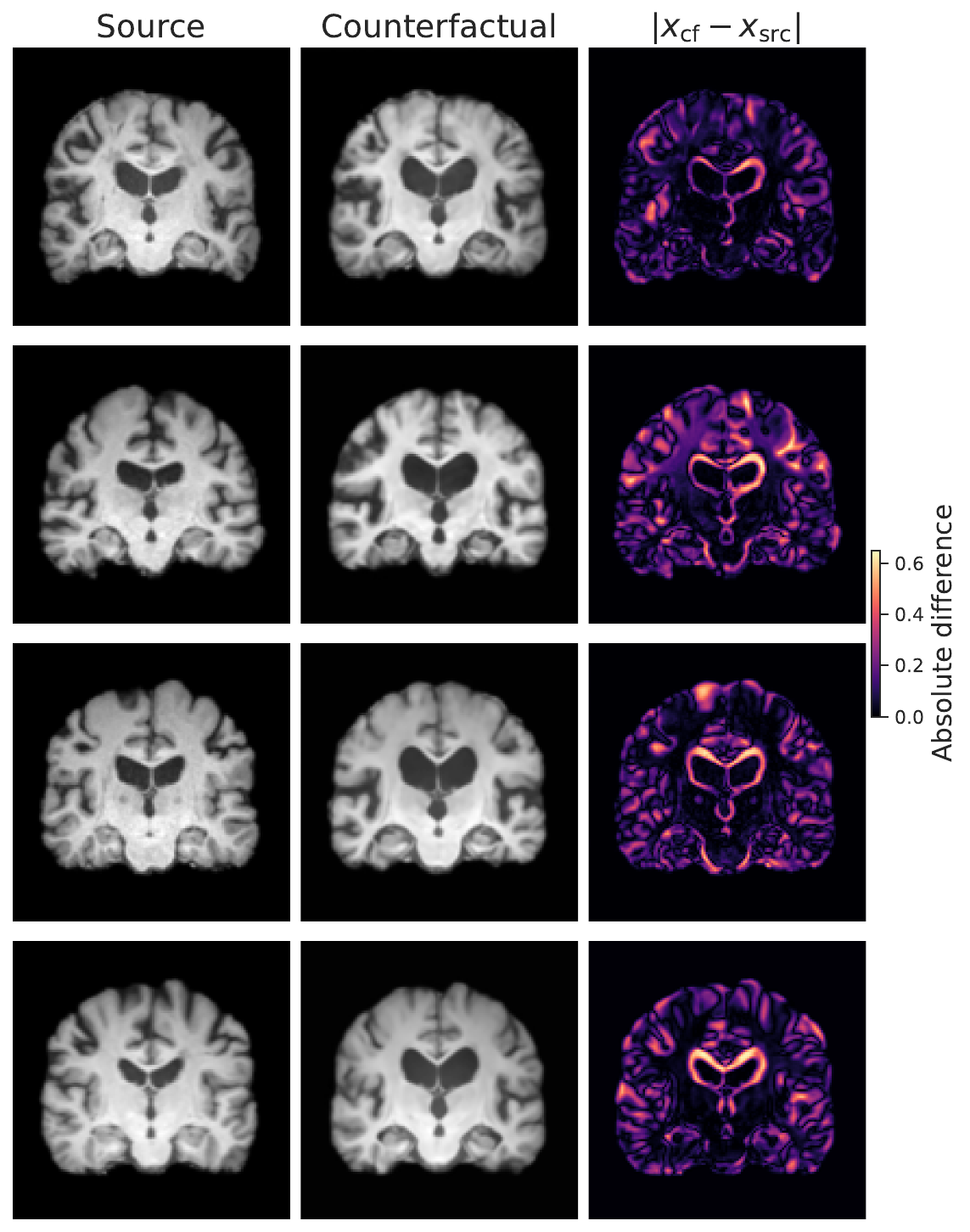}
        \vspace{2mm}
        \small (a) CDR
    \end{minipage}
    \hfill
    \begin{minipage}[t]{0.49\textwidth}
        \centering
        \includegraphics[width=\linewidth]{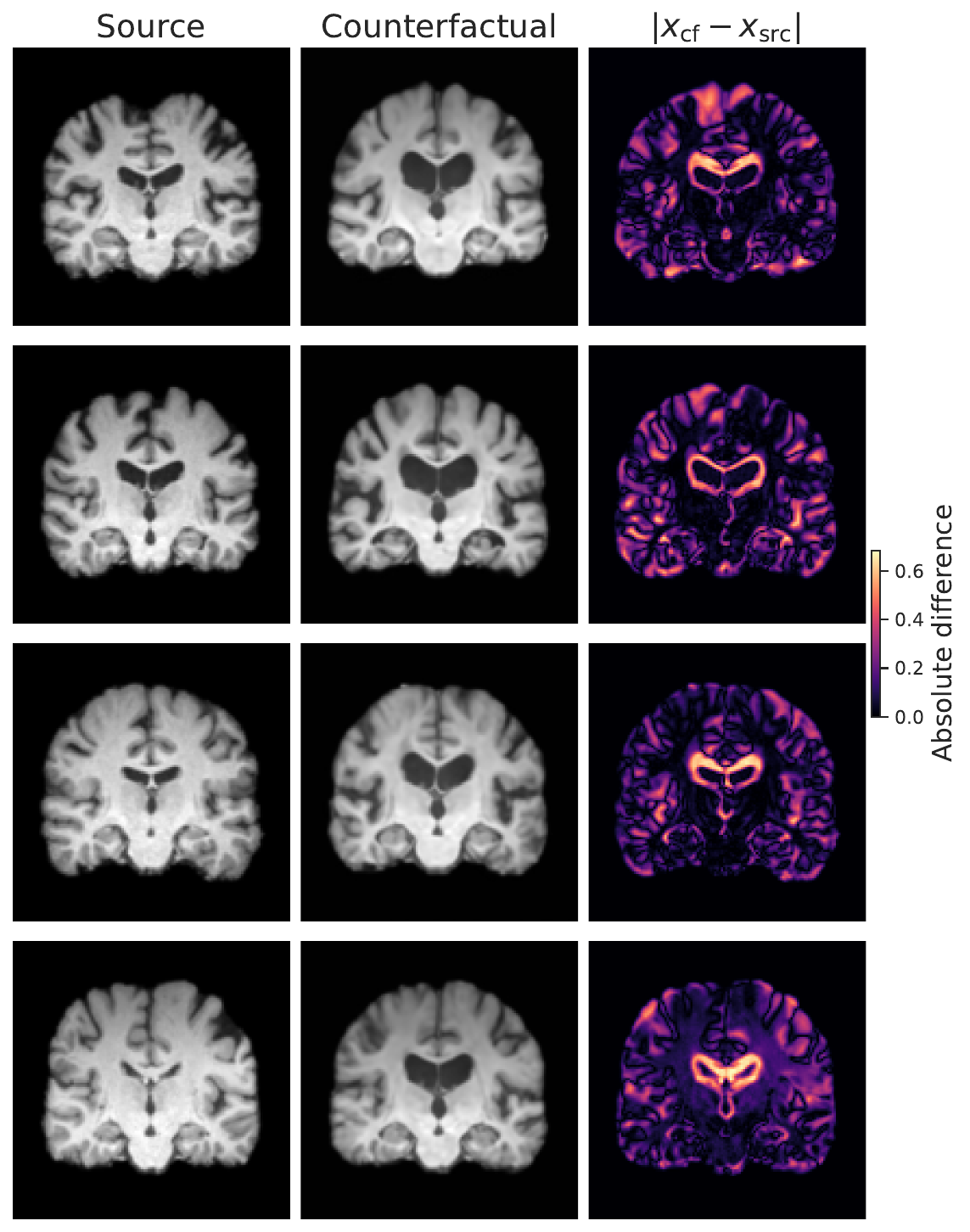}
        \vspace{2mm}
        \small (b) Age
    \end{minipage}
    \vspace{-5mm}
    \caption{Qualitative counterfactual examples generated from the proposed method for the MRI experiments. Each panel shows source images, generated counterfactuals, and absolute difference heatmaps. The heatmaps highlight the spatially localized edits induced by the counterfactual optimization. On the left column (a), the counterfactual generation is optimized by the CDR criteria separation, compared to the counterfactual generation optimized by the Age separation on the right column (b).}
    \label{fig:mri_cf_side_by_side}
\end{figure*}
Fig.~\ref{fig:mri_cf_side_by_side}(a) shows counterfactuals generated from healthy brains toward the cognitively impaired cohort (CDR $\geq 1$). The edited MRIs exhibit ventricular enlargement, increased CSF space, and more pronounced cortical sulci, which are consistent with anatomically more impaired brains. Fig.~\ref{fig:mri_cf_side_by_side}(b) shows counterfactuals generated from younger brains toward the older cohort (Age $> 74$), illustrating age-related structural changes under the same framework.
\subsection{Synthetic dSprites}

We evaluate the proposed method on dSprites \cite{dsprites17}, a synthetic dataset of 2D shapes generated from multiple latent factors. As an auxiliary dataset, we use Spaceshapes \cite{spaceshapes}, which contains three additional shape classes (moon, star, and ship). In this setting, Spaceshapes is used to train the auxiliary deep two-sample model.

\subsubsection{Experimental setup}

We consider a shape-transfer setting in which the source group consists of squares and the target group consists of ellipses. The auxiliary statistical model $f(\cdot)$ is implemented with a ResNet18 backbone and trained on Spaceshapes to classify the three shape categories. As in the MRI experiments, the pretrained deep two-sample test detects significant group differences on dSprites ($p < 0.05$). We evaluate counterfactual quality along three complementary axes: statistical effectiveness, minimality, and realism. Statistical effectiveness is measured by changes in the downstream two-sample test statistic and $p$-value, minimality is measured by LPIPS between source and counterfactual images, and realism is measured by PRDC between generated counterfactuals and target samples.
\subsubsection{Experimental Results}

\paragraph{Statistical effectiveness}
\begin{table}[t]
\centering
\caption{Main statistical evaluation on dSprites (source: square, target: ellipse) at $\lambda=4$. Larger $\Delta p$ and $\Delta t$ indicate more favorable shifts after counterfactual generation. ``Non-scale'' indicates that the reported test statistic is not normalized by the number of samples; see Eq.~\eqref{eq:dmmd}.}
\label{tab:dsprites_stats_lambda4}
\begin{tabular}{lccc}
\toprule
Method & $\Delta p$ mean & $\Delta t$ mean (non-scale) \\
\midrule
Conditional CF & \textbf{0.0352} & \textbf{0.2348}\\
Proposed CF & 0.0114 & 0.2081\\
\bottomrule
\end{tabular}
\end{table}
Table~\ref{tab:dsprites_stats_lambda4} reports the main statistical comparison at $\lambda=4$, matching the MRI setting. The proposed method yields positive mean changes in both $p$-value and test statistic, indicating that the generated counterfactuals move the source samples toward the target distribution under the downstream two-sample test. However, unlike in the MRI experiments, the proposed method does not outperform the conditional baseline on the primary statistical criterion in this setting. The gap becomes smaller as $\lambda$ increases, but the baseline remains stronger at the selected operating point.

\paragraph{Minimality and Realism}

\begin{table}[t]
\centering
\caption{Minimality evaluation on dSprites at $\lambda=4$ using LPIPS between source and counterfactual images. Lower is better.}
\label{tab:dsprites_lpips_lambda4}
\begin{tabular}{lc}
\toprule
Method & LPIPS \\
\midrule
Conditional CF & 0.1124 \\
Proposed CF & \textbf{0.0925} \\
\bottomrule
\end{tabular}
\end{table}

Table~\ref{tab:dsprites_lpips_lambda4} reports minimality at $\lambda=4$. Here, the proposed method outperforms the conditional baseline, achieving lower LPIPS and therefore producing edits that remain perceptually closer to the source squares while still moving toward the target ellipse class.
\begin{table}[t]
\centering
\caption{Realism evaluation on dSprites at $\lambda=4$ using PRDC metrics between generated counterfactuals and target samples. Higher is better.}
\label{tab:dsprites_prdc}
\begin{tabular}{lcccc}
\toprule
Method & Density & Coverage & Precision & Recall \\
\midrule
Conditional CF & 0.5227 & 0.4394 & 0.6317 & 0.7921 \\
Proposed CF & 0.1987 & 0.1986 & 0.2517 & 0.7758 \\
\bottomrule
\end{tabular}
\end{table}

To complement the statistical and minimality analyses, we also evaluate realism with PRDC. On dSprites, the conditional baseline is consistently stronger on realism, achieving higher density, coverage, precision, and recall. This is expected, since the baseline is optimized directly for conditional generation toward the target class, whereas the proposed method is optimized for reducing discrepancy under the deep two-sample test. Taken together, the dSprites results highlight a trade-off: the baseline is stronger on realism and on the primary statistical criterion, while the proposed method yields more minimal and source-faithful edits.
\paragraph{Qualitative interpretation}
\begin{figure*}[t]
    \centering
    \begin{minipage}[t]{\textwidth}
        \centering
        \includegraphics[width=\linewidth]{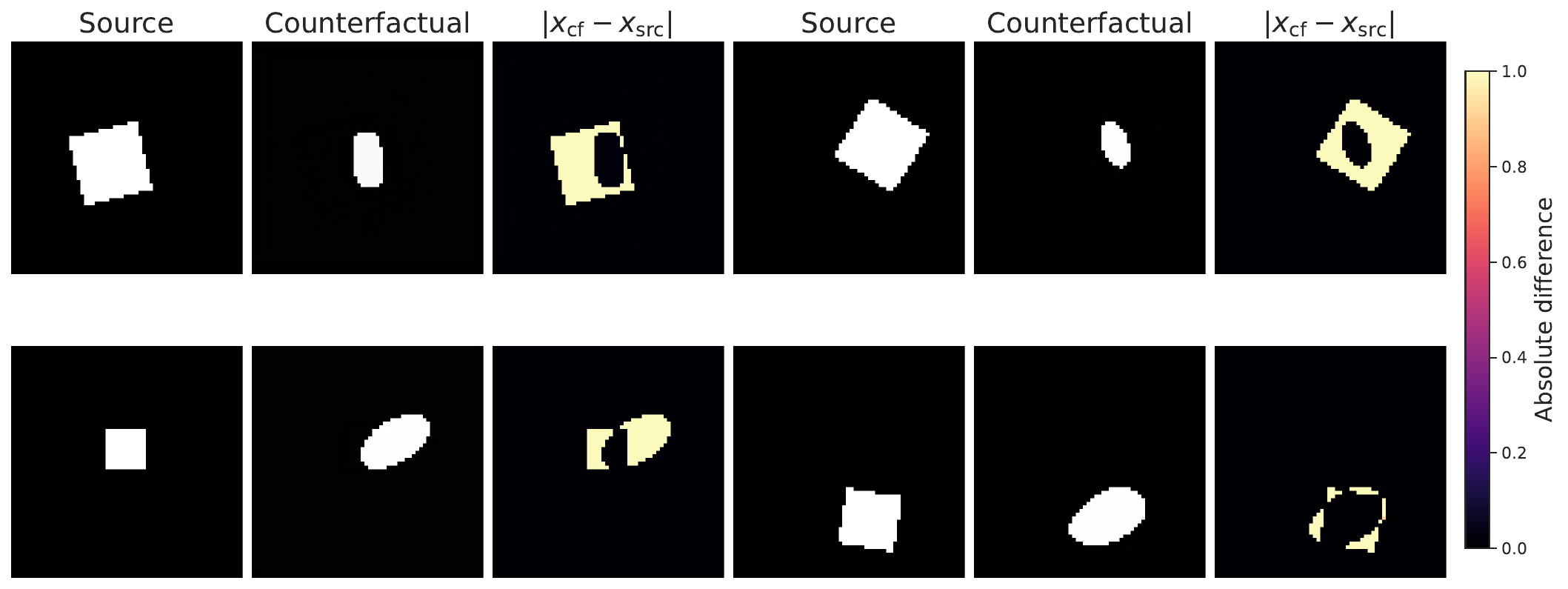}
    \end{minipage}
    \caption{Qualitative counterfactual examples generated from the proposed method for the dSprites shape experiments. Each panel shows source images, generated counterfactuals, and absolute difference heatmaps. The heatmaps highlight the spatially localized edits induced by the counterfactual optimization. Generated counterfactuals move toward the target group (ellipse).}
    \label{fig:dsprites_cf_shape}
\end{figure*}

Fig.~\ref{fig:dsprites_cf_shape} shows qualitative counterfactual examples for the dSprites setting, where source samples are squares and the target group consists of ellipses. The method generates edits that move the source shapes toward the target class.

\section{Ablation study over $\lambda$ and the number of edited source samples}
\begin{figure*}[t]
    \centering
    \begin{minipage}[t]{\textwidth}
        \centering
        \includegraphics[width=\linewidth]{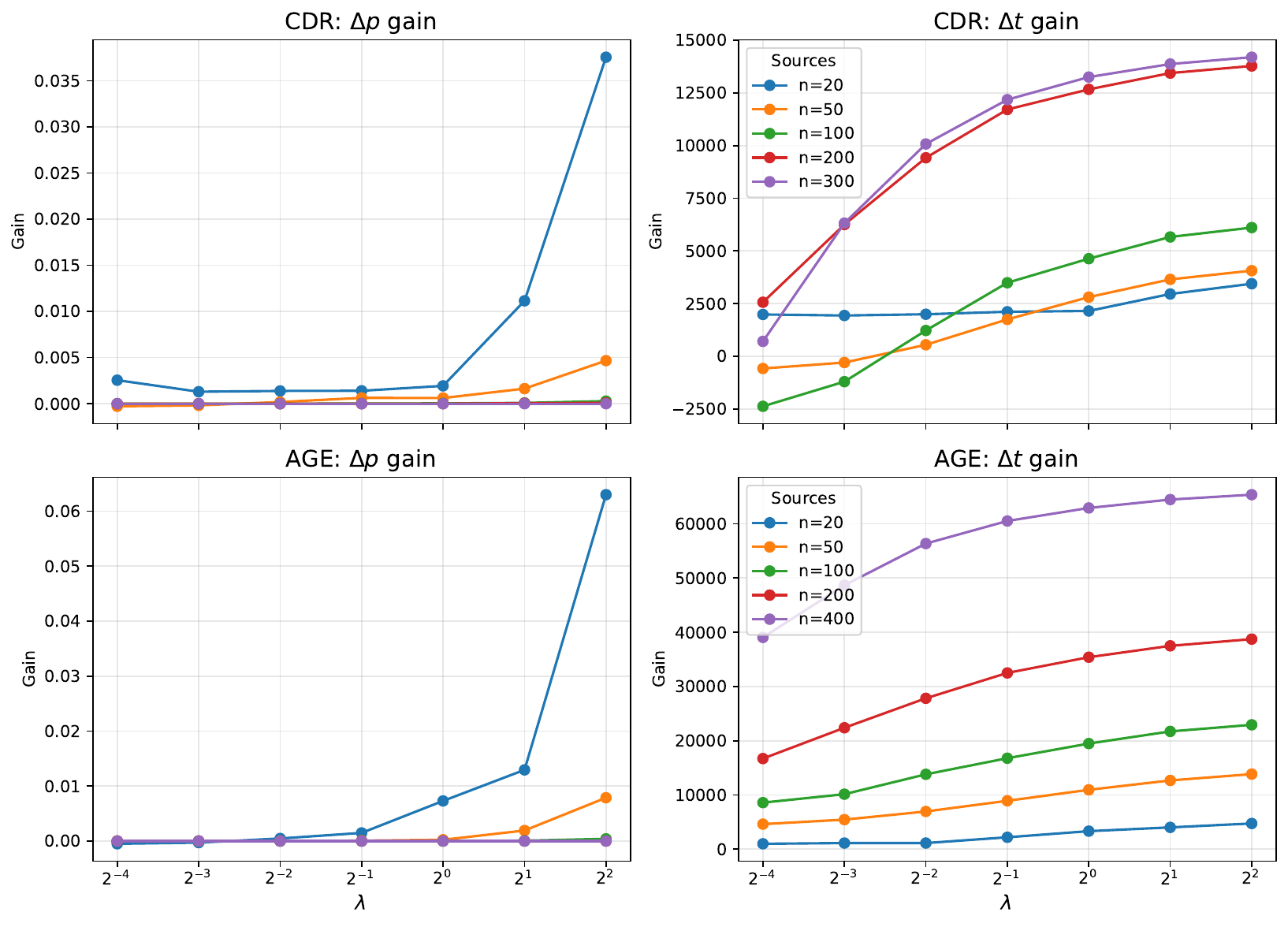}
    \end{minipage}
    \caption{Ablation study over the regularization weight $\lambda$ and the number of edited source samples $n$ in the MRI experiments. Top row: CDR-based source/target separation. Bottom row: Age-based source/target separation. Left column:
gain in downstream two-sample p-value ($\Delta p$). Right column: gain in discrepancy reduction ($\Delta t$). Larger values are better in all panels.
Across both criteria, increasing $\lambda$ consistently improves $\Delta t$, while gains in $\Delta p$ are most pronounced in the low-source regime.}
    \label{fig:ablation_lambda_nsource}
\end{figure*}
We study the sensitivity of the proposed method to the regularization 
$\lambda$ for LPIPS and to the number of edited source samples $n$, using the same statistical criteria as in the main experiments: the increase in downstream two-sample p-value ($\Delta p$) and the reduction in the discrepancy test ($\Delta t$), where larger values indicate that the edited source set is statistically closer to the target set. Figure~\ref{fig:ablation_lambda_nsource} summarizes these trends for the MRI experiments under both CDR- and Age-based cohort definitions.

For \textbf{CDR}, increasing $\lambda$ consistently improves $\Delta t$ across all tested source counts. The gain in $\Delta p$ is most visible in the low-source regime, especially for $n=20$ and $n=50$, whereas for larger source sets the pre-edit p-values are already close to the numerical floor (close to 0) and $\Delta p$ becomes less informative. In contrast, $\Delta t$ continues to increase with $\lambda$, indicating that stronger regularization still improves alignment with the target distribution even when p-value gains saturate. 

For \textbf{Age}, we observe the same overall pattern, but with clearer gains in the low-source regime. In particular, for small $n$, increasing $\lambda$ substantially improves both $\Delta p$ and $\Delta t$. As in the CDR setting, the p-value gains become smaller for larger source sets because the initial
p-values are already near zero, while $\Delta t$ continues to show a strong and stable improvement with increasing $\lambda$. Overall, the ablation indicates that $\lambda$ is the main driver of statistical improvement, while increasing
the number of edited source samples mainly stabilizes the effect-size metric $\Delta t$. 
\section{Discussion and Conclusion}

We introduced a counterfactual generation framework guided by a deep two-sample test to enable sample-level interpretation of distribution differences. To our knowledge, this is among the first approaches to use the discrepancy learned by a deep two-sample test as a direct supervision signal for generating counterfactual explanations.

Empirically, the proposed method is most convincing in the MRI experiments, where it improves the primary statistical criteria relative to the conditional baseline and produces anatomically meaningful counterfactuals. The ablation further suggests that the LPIPS regularization weight $\lambda$ helps control the trade-off between distribution alignment and source similarity, with the best statistical performance observed at $\lambda=4$. On dSprites, the method improves minimality, but its advantages on the main statistical objective are weaker.

Overall, our findings suggest that deep-test-guided counterfactual editing is a promising direction for interpretable analysis of distribution shifts. More broadly, they indicate that counterfactual explanations can be tailored not only to predictive models but also to statistical testing procedures, which may be especially useful in scientific and biomedical applications where understanding group differences is a primary objective.

\clearpage 

\acks{The authors gratefully acknowledge funding by the Deutsche Forschungsgemeinschaft (DFG, German Research Foundation) – 459422098. Furthermore, the authors thank the participants of the UK Biobank study and the ADNI study. This research has been conducted using the UK Biobank Resource under Application Number 77717. Furthermore,
data used in preparation of this article were obtained from the Alzheimer's Disease Neuroimaging Initiative (ADNI) database (adni.loni.usc.edu). As such, the investigators within the ADNI contributed to the design and implementation of ADNI and/or provided data but did not participate in the analysis or writing of this report. A complete listing of ADNI investigators can be found at: \url{http://adni.loni.usc.edu/wp-content/uploads/how_to_apply/ADNI_Acknowledgement_List.pdf}

}

\paragraph{Data availability \\}
\hspace{-5mm}
The MR imaging is available by applications from the UK Biobank (\url{https://www.ukbiobank.ac.uk/register-apply}) and the ADNI database (\url{https://adni.loni.usc.edu/data-samples/access-data/}).

\vskip 0.2in
\bibliography{sample}

@article{UKB,
   author = {Cathie Sudlow and John Gallacher and Naomi Allen and Valerie Beral and Paul Burton and John Danesh and Paul Downey and Paul Elliott and Jane Green and Martin Landray and Bette Liu and Paul Matthews and Giok Ong and Jill Pell and Alan Silman and Alan Young and Tim Sprosen and Tim Peakman and Rory Collins},
   doi = {10.1371/JOURNAL.PMED.1001779},
   issn = {15491676},
   issue = {3},
   journal = {PLoS Medicine},
   month = {3},
   pmid = {25826379},
   publisher = {Public Library of Science},
   title = {UK Biobank: An Open Access Resource for Identifying the Causes of a Wide Range of Complex Diseases of Middle and Old Age},
   volume = {12},
   year = {2015},
}

@ARTICLE{adni_Petersen2010-zh,
  title     = "Alzheimer's Disease Neuroimaging Initiative ({ADNI)}: clinical
               characterization",
  author    = "Petersen, R C and Aisen, P S and Beckett, L A and Donohue, M C
               and Gamst, A C and Harvey, D J and Jack, Jr, C R and Jagust, W J
               and Shaw, L M and Toga, A W and Trojanowski, J Q and Weiner, M W",
  journal   = "Neurology",
  publisher = "American Academy of Neurology",
  volume    =  74,
  number    =  3,
  pages     = "201--209",
  month     =  jan,
  year      =  2010
}

@article{ernst2004permutation,
  title={Permutation methods: a basis for exact inference},
  author={Ernst, Michael D},
  journal={Statistical Science},
  pages={676--685},
  year={2004},
  publisher={JSTOR}
}

@article{JMLR:v13:gretton12a,
  author  = {Arthur Gretton and Karsten M. Borgwardt and Malte J. Rasch and Bernhard Sch{{\"o}}lkopf and Alexander Smola},
  title   = {A Kernel Two-Sample Test},
  journal = {Journal of Machine Learning Research},
  year    = {2012},
  volume  = {13},
  number  = {25},
  pages   = {723-773},
  url     = {http://jmlr.org/papers/v13/gretton12a.html}
}

@inproceedings{Kirchler2020, 
title={Two-sample Testing Using Deep Learning}, 
ISSN={2640-3498}, url={https://proceedings.mlr.press/v108/kirchler20a.html}, 
abstractNote={We propose a two-sample testing procedure based on learned deep neural network representations. To this end, we define two test statistics that perform an asymptotic location test on data samples mapped onto a hidden layer. The tests are consistent and asymptotically control the type-1 error rate. Their test statistics can be evaluated in linear time (in the sample size). Suitable data representations are obtained in a data-driven way, by solving a supervised or unsupervised transfer-learning task on an auxiliary (potentially distinct) data set. If no auxiliary data is available, we split the data into two chunks: one for learning representations and one for computing the test statistic. In experiments on audio samples, natural images and three-dimensional neuroimaging data our tests yield significant decreases in type-2 error rate (up to 35 percentage points) compared to state-of-the-art two-sample tests such as kernel-methods and classifier two-sample tests.}, booktitle={Proceedings of the Twenty Third International Conference on Artificial Intelligence and Statistics}, 
publisher={PMLR}, 
author={Kirchler, Matthias and Khorasani, Shahryar and Kloft, Marius and Lippert, Christoph}, year={2020}, month=june, pages={1387–1398}, language={en}}

@inproceedings{globalCF, 
author = {Sobieski, Bartlomiej and Biecek, Przemyslaw}, title = {Global Counterfactual Directions}, year = {2024}, isbn = {978-3-031-73035-1}, publisher = {Springer-Verlag}, address = {Berlin, Heidelberg}, 
doi = {10.1007/978-3-031-73036-8_5}, booktitle = {Computer Vision – ECCV 2024: 18th European Conference, Milan, Italy, September 29–October 4, 2024, Proceedings, Part LXIII}, pages = {72–90}, numpages = {19}, location = {Milan, Italy} }

@inproceedings{preechakul2021diffusion,
      title={Diffusion Autoencoders: Toward a Meaningful and Decodable Representation}, 
      author={Preechakul, Konpat and Chatthee, Nattanat and Wizadwongsa, Suttisak and Suwajanakorn, Supasorn},
      booktitle={IEEE Conference on Computer Vision and Pattern Recognition (CVPR)}, 
      year={2022},
}

@inproceedings{liu2020learning,
  title={Learning deep kernels for non-parametric two-sample tests},
  author={Liu, Feng and Xu, Wenkai and Lu, Jie and Zhang, Guangquan and Gretton, Arthur and Sutherland, Danica J},
  booktitle={International conference on machine learning},
  pages={6316--6326},
  year={2020},
  organization={PMLR}
}

@inproceedings{
lopez-paz2017revisiting,
title={Revisiting Classifier Two-Sample Tests},
author={David Lopez-Paz and Maxime Oquab},
booktitle={International Conference on Learning Representations},
year={2017},
url={https://openreview.net/forum?id=SJkXfE5xx}
}

@INPROCEEDINGS{gradcam,
  author={Selvaraju, Ramprasaath R. and Cogswell, Michael and Das, Abhishek and Vedantam, Ramakrishna and Parikh, Devi and Batra, Dhruv},
  booktitle={2017 IEEE International Conference on Computer Vision (ICCV)}, 
  title={Grad-CAM: Visual Explanations from Deep Networks via Gradient-Based Localization}, 
  year={2017},
  volume={},
  number={},
  pages={618-626},
  doi={10.1109/ICCV.2017.74}}

@inproceedings{grad_input, author = {Shrikumar, Avanti and Greenside, Peyton and Kundaje, Anshul}, title = {Learning important features through propagating activation differences}, year = {2017}, publisher = {JMLR.org}, booktitle = {Proceedings of the 34th International Conference on Machine Learning - Volume 70}, pages = {3145–3153}, numpages = {9}, location = {Sydney, NSW, Australia}, series = {ICML'17} }

@InProceedings{guided_backprop,
  author       = "J.T. Springenberg and A. Dosovitskiy and T. Brox and M. Riedmiller",
  title        = "Striving for Simplicity: The All Convolutional Net",
  booktitle    = "ICLR (workshop track)",
  year         = "2015",
  url          = "http://lmb.informatik.uni-freiburg.de/Publications/2015/DB15a"
}

@inproceedings{integrated_grad,
author = {Sundararajan, Mukund and Taly, Ankur and Yan, Qiqi},
title = {Axiomatic attribution for deep networks},
year = {2017},
publisher = {JMLR.org},
booktitle = {Proceedings of the 34th International Conference on Machine Learning - Volume 70},
pages = {3319–3328},
numpages = {10},
location = {Sydney, NSW, Australia},
series = {ICML'17}
}

@InProceedings{gifsplanation_cohen,
  title = 	 {Gifsplanation via Latent Shift: A Simple Autoencoder Approach to Counterfactual Generation for Chest X-rays},
  author =       {Cohen, Joseph Paul and Brooks, Rupert and En, Sovann and Zucker, Evan and Pareek, Anuj and Lungren, Matthew P. and Chaudhari, Akshay},
  booktitle = 	 {Proceedings of the Fourth Conference on Medical Imaging with Deep Learning},
  pages = 	 {74--104},
  year = 	 {2021},
  editor = 	 {Heinrich, Mattias and Dou, Qi and de Bruijne, Marleen and Lellmann, Jan and Schläfer, Alexander and Ernst, Floris},
  volume = 	 {143},
  series = 	 {Proceedings of Machine Learning Research},
  month = 	 {07--09 Jul},
  publisher =    {PMLR},
  url = 	 {https://proceedings.mlr.press/v143/cohen21a.html}
}

@article{counterfactual_singla,
title = {Explaining the black-box smoothly—A counterfactual approach},
journal = {Medical Image Analysis},
volume = {84},
pages = {102721},
year = {2023},
issn = {1361-8415},
doi = {https://doi.org/10.1016/j.media.2022.102721},
url = {https://www.sciencedirect.com/science/article/pii/S1361841522003498},
author = {Sumedha Singla and Motahhare Eslami and Brian Pollack and Stephen Wallace and Kayhan Batmanghelich}
}

@inproceedings{
Singla2020Explanation,
title={Explanation  by Progressive  Exaggeration},
author={Sumedha Singla and Brian Pollack and Junxiang Chen and Kayhan Batmanghelich},
booktitle={International Conference on Learning Representations},
year={2020},
url={https://openreview.net/forum?id=H1xFWgrFPS}
}

@article{tian2024unified,
  title={A unified data representation learning for non-parametric two-sample testing},
  author={Tian, Xunye and Peng, Liuhua and Zhou, Zhijian and Gong, Mingming and Gretton, Arthur and Liu, Feng},
  journal={arXiv preprint arXiv:2412.00613},
  year={2024}
}

@inproceedings{zhang2018_lpips,
  title={The Unreasonable Effectiveness of Deep Features as a Perceptual Metric},
  author={Zhang, Richard and Isola, Phillip and Efros, Alexei A and Shechtman, Eli and Wang, Oliver},
  booktitle={CVPR},
  year={2018}
}

@inproceedings{ho_ddpm,
author = {Ho, Jonathan and Jain, Ajay and Abbeel, Pieter},
title = {Denoising diffusion probabilistic models},
year = {2020},
isbn = {9781713829546},
publisher = {Curran Associates Inc.},
address = {Red Hook, NY, USA},
booktitle = {Proceedings of the 34th International Conference on Neural Information Processing Systems},
articleno = {574},
numpages = {12},
location = {Vancouver, BC, Canada},
series = {NIPS '20}
}

@InProceedings{improved_ddpm-nichol21a,
  title = 	 {Improved Denoising Diffusion Probabilistic Models},
  author =       {Nichol, Alexander Quinn and Dhariwal, Prafulla},
  booktitle = 	 {Proceedings of the 38th International Conference on Machine Learning},
  pages = 	 {8162--8171},
  year = 	 {2021},
  editor = 	 {Meila, Marina and Zhang, Tong},
  volume = 	 {139},
  series = 	 {Proceedings of Machine Learning Research},
  month = 	 {18--24 Jul},
  publisher =    {PMLR},
  url = 	 {https://proceedings.mlr.press/v139/nichol21a.html}
}

@InProceedings{Rombach_2022_ldm,
    author    = {Rombach, Robin and Blattmann, Andreas and Lorenz, Dominik and Esser, Patrick and Ommer, Bj\"orn},
    title     = {High-Resolution Image Synthesis With Latent Diffusion Models},
    booktitle = {Proceedings of the IEEE/CVF Conference on Computer Vision and Pattern Recognition (CVPR)},
    month     = {June},
    year      = {2022},
    pages     = {10684-10695}
}

@inproceedings{KingmaW13_vae,
  author       = {Diederik P. Kingma and
                  Max Welling},
  editor       = {Yoshua Bengio and
                  Yann LeCun},
  title        = {Auto-Encoding Variational Bayes},
  booktitle    = {2nd International Conference on Learning Representations, {ICLR} 2014,
                  Banff, AB, Canada, April 14-16, 2014, Conference Track Proceedings},
  year         = {2014},
  url          = {http://arxiv.org/abs/1312.6114},
  bibsource    = {dblp computer science bibliography, https://dblp.org}
}

@inproceedings{NIPS2014_gan,
 author = {Goodfellow, Ian J. and Pouget-Abadie, Jean and Mirza, Mehdi and Xu, Bing and Warde-Farley, David and Ozair, Sherjil and Courville, Aaron and Bengio, Yoshua},
 booktitle = {Advances in Neural Information Processing Systems},
 editor = {Z. Ghahramani and M. Welling and C. Cortes and N. Lawrence and K.Q. Weinberger},
 pages = {},
 publisher = {Curran Associates, Inc.},
 title = {Generative Adversarial Nets},
 volume = {27},
 year = {2014}
}

@article{xai_medical,
title = {Explainable AI in medical imaging: An overview for clinical practitioners – Beyond saliency-based XAI approaches},
journal = {European Journal of Radiology},
volume = {162},
pages = {110786},
year = {2023},
issn = {0720-048X},
doi = {https://doi.org/10.1016/j.ejrad.2023.110786},
url = {https://www.sciencedirect.com/science/article/pii/S0720048X23001006},
author = {Katarzyna Borys and Yasmin Alyssa Schmitt and Meike Nauta and Christin Seifert and Nicole Krämer and Christoph M. Friedrich and Felix Nensa}
}

@InProceedings{spaceshapes,
  title = 	 {Benchmarks, Algorithms, and Metrics for Hierarchical Disentanglement},
  author =       {Ross, Andrew and Doshi-Velez, Finale},
  booktitle = 	 {Proceedings of the 38th International Conference on Machine Learning},
  pages = 	 {9084--9094},
  year = 	 {2021},
  editor = 	 {Meila, Marina and Zhang, Tong},
  volume = 	 {139},
  series = 	 {Proceedings of Machine Learning Research},
  month = 	 {18--24 Jul},
  publisher =    {PMLR},
  url = 	 {https://proceedings.mlr.press/v139/ross21a.html}
}

@misc{dsprites17,
author = {Loic Matthey and Irina Higgins and Demis Hassabis and Alexander Lerchner},
title = {dSprites: Disentanglement testing Sprites dataset},
howpublished= {https://github.com/deepmind/dsprites-dataset/},
year = "2017",
}

\end{document}